# Convolutional Sequence to Sequence Learning

**Jonas Gehring**
**Michael Auli**
**David Grangier**
**Denis Yarats**
**Yann N. Dauphin**
Facebook AI Research

## Abstract

The prevalent approach to sequence to sequence learning maps an input sequence to a variable length output sequence via recurrent neural networks. We introduce an architecture based entirely on convolutional neural networks.[1] Compared to recurrent models, computations over all elements can be fully parallelized during training to better exploit the GPU hardware and optimization is easier since the number of non-linearities is fixed and independent of the input length. Our use of gated linear units eases gradient propagation and we equip each decoder layer with a separate attention module. We outperform the accuracy of the deep LSTM setup of Wu et al. (2016) on both WMT'14 English-German and WMT'14 English-French translation at an order of magnitude faster speed, both on GPU and CPU.

## 1. Introduction

Sequence to sequence learning has been successful in many tasks such as machine translation, speech recognition (Sutskever et al., 2014; Chorowski et al., 2015) and text summarization (Rush et al., 2015; Nallapati et al., 2016; Shen et al., 2016) amongst others. The dominant approach to date encodes the input sequence with a series of bi-directional recurrent neural networks (RNN) and generates a variable length output with another set of decoder RNNs, both of which interface via a soft-attention mechanism (Bahdanau et al., 2014; Luong et al., 2015). In machine translation, this architecture has been demonstrated to outperform traditional phrase-based models by large margins (Sennrich et al., 2016b; Zhou et al., 2016; Wu et al., 2016; §2).

---

[1]The source code and models are available at https://github.com/facebookresearch/fairseq.

Convolutional neural networks are less common for sequence modeling, despite several advantages (Waibel et al., 1989; LeCun & Bengio, 1995). Compared to recurrent layers, convolutions create representations for fixed size contexts, however, the effective context size of the network can easily be made larger by stacking several layers on top of each other. This allows to precisely control the maximum length of dependencies to be modeled. Convolutional networks do not depend on the computations of the previous time step and therefore allow parallelization over every element in a sequence. This contrasts with RNNs which maintain a hidden state of the entire past that prevents parallel computation within a sequence.

Multi-layer convolutional neural networks create hierarchical representations over the input sequence in which nearby input elements interact at lower layers while distant elements interact at higher layers. Hierarchical structure provides a shorter path to capture long-range dependencies compared to the chain structure modeled by recurrent networks, e.g. we can obtain a feature representation capturing relationships within a window of $n$ words by applying only $\mathcal{O}(\frac{n}{k})$ convolutional operations for kernels of width $k$, compared to a linear number $\mathcal{O}(n)$ for recurrent neural networks. Inputs to a convolutional network are fed through a constant number of kernels and non-linearities, whereas recurrent networks apply up to $n$ operations and non-linearities to the first word and only a single set of operations to the last word. Fixing the number of non-linearities applied to the inputs also eases learning.

Recent work has applied convolutional neural networks to sequence modeling such as Bradbury et al. (2016) who introduce recurrent pooling between a succession of convolutional layers or Kalchbrenner et al. (2016) who tackle neural translation without attention. However, none of these approaches has been demonstrated improvements over state of the art results on large benchmark datasets. Gated convolutions have been previously explored for machine translation by Meng et al. (2015) but their evaluation was restricted to a small dataset and the model was used in tandem with a traditional count-based model. Architec-tures which are partially convolutional have shown strong performance on larger tasks but their decoder is still recurrent (Gehring et al., 2016).

In this paper we propose an architecture for sequence to sequence modeling that is entirely convolutional. Our model is equipped with gated linear units (Dauphin et al., 2016) and residual connections (He et al., 2015a). We also use attention in every decoder layer and demonstrate that each attention layer only adds a negligible amount of overhead. The combination of these choices enables us to tackle large scale problems (§3).

We evaluate our approach on several large datasets for machine translation as well as summarization and compare to the current best architectures reported in the literature. On WMT'16 English-Romanian translation we achieve a new state of the art, outperforming the previous best result by 1.9 BLEU. On WMT'14 English-German we outperform the strong LSTM setup of Wu et al. (2016) by 0.5 BLEU and on WMT'14 English-French we outperform the likelihood trained system of Wu et al. (2016) by 1.6 BLEU. Furthermore, our model can translate unseen sentences at an order of magnitude faster speed than Wu et al. (2016) on GPU and CPU hardware (§4, §5).

## 2. Recurrent Sequence to Sequence Learning

Sequence to sequence modeling has been synonymous with recurrent neural network based encoder-decoder architectures (Sutskever et al., 2014; Bahdanau et al., 2014). The encoder RNN processes an input sequence $\mathbf{x} = (x_1, \ldots, x_m)$ of $m$ elements and returns state representations $\mathbf{z} = (z_1, \ldots, z_m)$. The decoder RNN takes $\mathbf{z}$ and generates the output sequence $\mathbf{y} = (y_1, \ldots, y_n)$ left to right, one element at a time. To generate output $y_{i+1}$, the decoder computes a new hidden state $h_{i+1}$ based on the previous state $h_i$, an embedding $g_i$ of the previous target language word $y_i$, as well as a conditional input $c_i$ derived from the encoder output $\mathbf{z}$. Based on this generic formulation, various encoder-decoder architectures have been proposed, which differ mainly in the conditional input and the type of RNN.

Models without attention consider only the final encoder state $z_m$ by setting $c_i = z_m$ for all $i$ (Cho et al., 2014), or simply initialize the first decoder state with $z_m$ (Sutskever et al., 2014), in which case $c_i$ is not used. Architectures with attention (Bahdanau et al., 2014; Luong et al., 2015) compute $c_i$ as a weighted sum of $(z_1, \ldots, z_m)$ at each time step. The weights of the sum are referred to as attention scores and allow the network to focus on different parts of the input sequence as it generates the output sequences. Attention scores are computed by essentially comparing each encoder state $z_j$ to a combination of the previous decoder state $h_i$ and the last prediction $y_i$; the result is normalized to be a distribution over input elements.

Popular choices for recurrent networks in encoder-decoder models are long short term memory networks (LSTM; Hochreiter & Schmidhuber, 1997) and gated recurrent units (GRU; Cho et al., 2014). Both extend Elman RNNs (Elman, 1990) with a gating mechanism that allows the memorization of information from previous time steps in order to model long-term dependencies. Most recent approaches also rely on bi-directional encoders to build representations of both past and future contexts (Bahdanau et al., 2014; Zhou et al., 2016; Wu et al., 2016). Models with many layers often rely on shortcut or residual connections (He et al., 2015a; Zhou et al., 2016; Wu et al., 2016).

## 3. A Convolutional Architecture

Next we introduce a fully convolutional architecture for sequence to sequence modeling. Instead of relying on RNNs to compute intermediate encoder states $\mathbf{z}$ and decoder states $\mathbf{h}$ we use convolutional neural networks (CNN).

### 3.1. Position Embeddings

First, we embed input elements $\mathbf{x} = (x_1, \ldots, x_m)$ in distributional space as $\mathbf{w} = (w_1, \ldots, w_m)$, where $w_j \in \mathbb{R}^f$ is a column in an embedding matrix $\mathcal{D} \in \mathbb{R}^{V \times f}$. We also equip our model with a sense of order by embedding the absolute position of input elements $\mathbf{p} = (p_1, \ldots, p_m)$ where $p_j \in \mathbb{R}^f$. Both are combined to obtain input element representations $\mathbf{e} = (w_1 + p_1, \ldots, w_m + p_m)$. We proceed similarly for output elements that were already generated by the decoder network to yield output element representations that are being fed back into the decoder network $\mathbf{g} = (g_1, \ldots, g_n)$. Position embeddings are useful in our architecture since they give our model a sense of which portion of the sequence in the input or output it is currently dealing with (§5.4).

### 3.2. Convolutional Block Structure

Both encoder and decoder networks share a simple block structure that computes intermediate states based on a fixed number of input elements. We denote the output of the $l$-th block as $\mathbf{h}^l = (h_1^l, \ldots, h_n^l)$ for the decoder network, and $\mathbf{z}^l = (z_1^l, \ldots, z_m^l)$ for the encoder network; we refer to blocks and layers interchangeably. Each block contains a one dimensional convolution followed by a non-linearity. For a decoder network with a single block and kernel width $k$, each resulting state $h_i^1$ contains information over $k$ input elements. Stacking several blocks on top of each other increases the number of input elements represented in a state. For instance, stacking 6 blocks with $k = 5$ results in an input field of 25 elements, i.e. each output depends on 25 inputs. Non-linearities allow the networks to exploit the full input field, or to focus on fewer elements if needed.

Each convolution kernel is parameterized as $W \in \mathbb{R}^{2d \times kd}$, $b_w \in \mathbb{R}^{2d}$ and takes as input $X \in \mathbb{R}^{k \times d}$ which is a concatenation of $k$ input elements embedded in $d$ dimensions and maps them to a single output element $Y \in \mathbb{R}^{2d}$ that has twice the dimensionality of the input elements; subsequent layers operate over the $k$ output elements of the previous layer. We choose gated linear units (GLU; Dauphin et al., 2016) as non-linearity which implement a simple gating mechanism over the output of the convolution $Y = [A \ B] \in \mathbb{R}^{2d}$:

$$v([A \ B]) = A \otimes \sigma(B)$$

where $A, B \in \mathbb{R}^d$ are the inputs to the non-linearity, $\otimes$ is the point-wise multiplication and the output $v([A \ B]) \in \mathbb{R}^d$ is half the size of $Y$. The gates $\sigma(B)$ control which inputs $A$ of the current context are relevant. A similar non-linearity has been introduced in Oord et al. (2016b) who apply tanh to $A$ but Dauphin et al. (2016) shows that GLUs perform better in the context of language modelling.

To enable deep convolutional networks, we add residual connections from the input of each convolution to the output of the block (He et al., 2015a).

$$h_i^l = v(W^l[h_{i-k/2}^{l-1}, \ldots, h_{i+k/2}^{l-1}] + b_w^l) + h_i^{l-1}$$

For encoder networks we ensure that the output of the convolutional layers matches the input length by padding the input at each layer. However, for decoder networks we have to take care that no future information is available to the decoder (Oord et al., 2016a). Specifically, we pad the input by $k - 1$ elements on both the left and right side by zero vectors, and then remove $k$ elements from the end of the convolution output.

We also add linear mappings to project between the embedding size $f$ and the convolution outputs that are of size $2d$. We apply such a transform to $\mathbf{w}$ when feeding embeddings to the encoder network, to the encoder output $z_j^u$, to the final layer of the decoder just before the softmax $\mathbf{h^L}$, and to all decoder layers $\mathbf{h^l}$ before computing attention scores (1).

Finally, we compute a distribution over the $T$ possible next target elements $y_{i+1}$ by transforming the top decoder output $h_i^L$ via a linear layer with weights $W_o$ and bias $b_o$:

$$p(y_{i+1}|y_1, \ldots, y_i, \mathbf{x}) = \text{softmax}(W_o h_i^L + b_o) \in \mathbb{R}^T$$

### 3.3. Multi-step Attention

We introduce a separate attention mechanism for each decoder layer. To compute the attention, we combine the current decoder state $h_i^l$ with an embedding of the previous

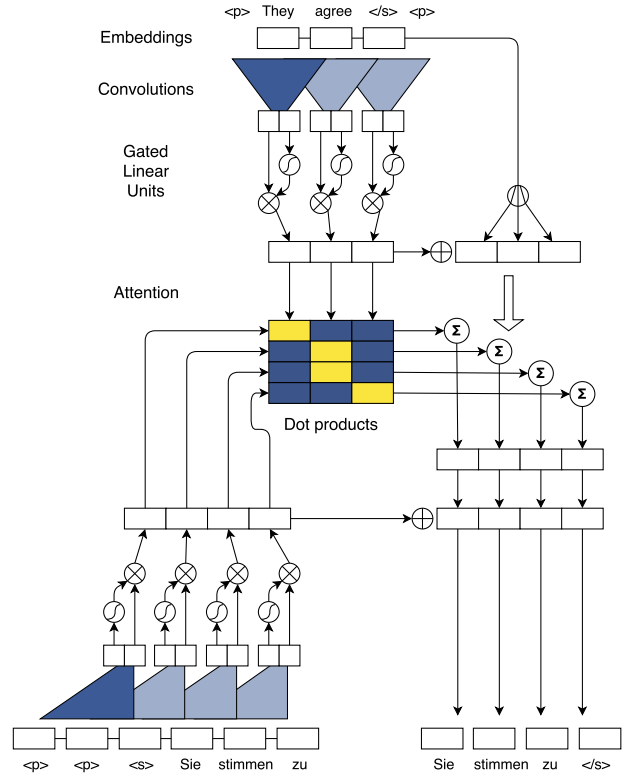

*Figure 1.* Illustration of batching during training. The English source sentence is encoded (top) and we compute all attention values for the four German target words (center) simultaneously. Our attentions are just dot products between decoder context representations (bottom left) and encoder representations. We add the conditional inputs computed by the attention (center right) to the decoder states which then predict the target words (bottom right). The sigmoid and multiplicative boxes illustrate Gated Linear Units.

target element $g_i$:

$$d_i^l = W_d^l h_i^l + b_d^l + g_i \qquad (1)$$

For decoder layer $l$ the attention $a_{ij}^l$ of state $i$ and source element $j$ is computed as a dot-product between the decoder state summary $d_i^l$ and each output $z_j^u$ of the last encoder block $u$:

$$a_{ij}^l = \frac{\exp\left(d_i^l \cdot z_j^u\right)}{\sum_{t=1}^m \exp\left(d_i^l \cdot z_t^u\right)}$$

The conditional input $c_i^l$ to the current decoder layer is a weighted sum of the encoder outputs as well as the input element embeddings $e_j$ (Figure 1, center right):

$$c_i^l = \sum_{j=1}^m a_{ij}^l(z_j^u + e_j) \qquad (2)$$

This is slightly different to recurrent approaches which compute both the attention and the weighted sum over $z_j^u$

only. We found adding $e_j$ to be beneficial and it resembles key-value memory networks where the keys are the $z_j^u$ and the values are the $z_j^u + e_j$ (Miller et al., 2016). Encoder outputs $z_j^u$ represent potentially large input contexts and $e_j$ provides point information about a specific input element that is useful when making a prediction. Once $c_i^l$ has been computed, it is simply added to the output of the corresponding decoder layer $h_i^l$.

This can be seen as attention with multiple 'hops' (Sukhbaatar et al., 2015) compared to single step attention (Bahdanau et al., 2014; Luong et al., 2015; Zhou et al., 2016; Wu et al., 2016). In particular, the attention of the first layer determines a useful source context which is then fed to the second layer that takes this information into account when computing attention etc. The decoder also has immediate access to the attention history of the $k - 1$ previous time steps because the conditional inputs $c_{i-k}^{l-1}, \ldots, c_i^{l-1}$ are part of $h_{i-k}^{l-1}, \ldots, h_i^{l-1}$ which are input to $h_i^l$. This makes it easier for the model to take into account which previous inputs have been attended to already compared to recurrent nets where this information is in the recurrent state and needs to survive several non-linearities. Overall, our attention mechanism considers which words we previously attended to (Yang et al., 2016) *and* performs multiple attention 'hops' per time step. In Appendix §C, we plot attention scores for a deep decoder and show that at different layers, different portions of the source are attended to.

Our convolutional architecture also allows to batch the attention computation across all elements of a sequence compared to RNNs (Figure 1, middle). We batch the computations of each decoder layer individually.

### 3.4. Normalization Strategy

We stabilize learning through careful weight initialization (§3.5) and by scaling parts of the network to ensure that the variance throughout the network does not change dramatically. In particular, we scale the output of residual blocks as well as the attention to preserve the variance of activations. We multiply the sum of the input and output of a residual block by $\sqrt{0.5}$ to halve the variance of the sum. This assumes that both summands have the same variance which is not always true but effective in practice.

The conditional input $c_i^l$ generated by the attention is a weighted sum of $m$ vectors (2) and we counteract a change in variance through scaling by $m\sqrt{1/m}$; we multiply by $m$ to scale up the inputs to their original size, assuming the attention scores are uniformly distributed. This is generally not the case but we found it to work well in practice.

For convolutional decoders with multiple attention, we scale the gradients for the encoder layers by the number of attention mechanisms we use; we exclude source word embeddings. We found this to stabilize learning since the encoder received too much gradient otherwise.

### 3.5. Initialization

Normalizing activations when adding the output of different layers, e.g. residual connections, requires careful weight initialization. The motivation for our initialization is the same as for the normalization: maintain the variance of activations throughout the forward and backward passes. All embeddings are initialized from a normal distribution with mean 0 and standard deviation 0.1. For layers whose output is not directly fed to a gated linear unit, we initialize weights from $\mathcal{N}(0, \sqrt{1/n_l})$ where $n_l$ is the number of input connections to each neuron. This ensures that the variance of a normally distributed input is retained.

For layers which are followed by a GLU activation, we propose a weight initialization scheme by adapting the derivations in (He et al., 2015b; Glorot & Bengio, 2010; Appendix A). If the GLU inputs are distributed with mean 0 and have sufficiently small variance, then we can approximate the output variance with $1/4$ of the input variance (Appendix A.1). Hence, we initialize the weights so that the input to the GLU activations have 4 times the variance of the layer input. This is achieved by drawing their initial values from $\mathcal{N}(0, \sqrt{4/n_l})$. Biases are uniformly set to zero when the network is constructed.

We apply dropout to the input of some layers so that inputs are retained with a probability of $p$. This can be seen as multiplication with a Bernoulli random variable taking value $1/p$ with probability $p$ and 0 otherwise (Srivastava et al., 2014). The application of dropout will then cause the variance to be scaled by $1/p$. We aim to restore the incoming variance by initializing the respective layers with larger weights. Specifically, we use $\mathcal{N}(0, \sqrt{4p/n_l})$ for layers whose output is subject to a GLU and $\mathcal{N}(0, \sqrt{p/n_l})$ otherwise (Appendix A.3).

## 4. Experimental Setup

### 4.1. Datasets

We consider three major WMT translation tasks as well as a text summarization task.

**WMT'16 English-Romanian.** We use the same data and pre-processing as Sennrich et al. (2016b) but remove sentences with more than 175 words. This results in 2.8M sentence pairs for training and we evaluate on newstest2016.[2]

---

[2] We followed the pre-processing of `https://github.com/rsennrich/wmt16-scripts/blob/80e21e5/sample/preprocess.sh` and added the back-translated data from `http://data.statmt.org/rsennrich/wmt16_`

We experiment with word-based models using a source vocabulary of 200K types and a target vocabulary of 80K types. We also consider a joint source and target byte-pair encoding (BPE) with 40K types (Sennrich et al., 2016a;b).

**WMT'14 English-German.** We use the same setup as Luong et al. (2015) which comprises 4.5M sentence pairs for training and we test on newstest2014.[3] As vocabulary we use 40K sub-word types based on BPE.

**WMT'14 English-French.** We use the full training set of 36M sentence pairs, and remove sentences longer than 175 words as well as pairs with a source/target length ratio exceeding 1.5. This results in 35.5M sentence-pairs for training. Results are reported on *newstest2014*. We use a source and target vocabulary with 40K BPE types.

In all setups a small subset of the training data serves as validation set (about 0.5-1% for each dataset) for early stopping and learning rate annealing.

**Abstractive summarization.** We train on the Gigaword corpus (Graff et al., 2003) and pre-process it identically to Rush et al. (2015) resulting in 3.8M training examples and 190K for validation. We evaluate on the DUC-2004 test data comprising 500 article-title pairs (Over et al., 2007) and report three variants of recall-based ROUGE (Lin, 2004), namely, ROUGE-1 (unigrams), ROUGE-2 (bigrams), and ROUGE-L (longest-common substring). We also evaluate on a Gigaword test set of 2000 pairs which is identical to the one used by Rush et al. (2015) and we report F1 ROUGE similar to prior work. Similar to Shen et al. (2016) we use a source and target vocabulary of 30K words and require outputs to be at least 14 words long.

### 4.2. Model Parameters and Optimization

We use 512 hidden units for both encoders and decoders, unless otherwise stated. All embeddings, including the output produced by the decoder before the final linear layer, have dimensionality 512; we use the same dimensionalities for linear layers mapping between the hidden and embedding sizes (§3.2).

We train our convolutional models with Nesterov's accelerated gradient method (Sutskever et al., 2013) using a momentum value of 0.99 and renormalize gradients if their norm exceeds 0.1 (Pascanu et al., 2013). We use a learning rate of 0.25 and once the validation perplexity stops improving, we reduce the learning rate by an order of magnitude after each epoch until it falls below $10^{-4}$.

Unless otherwise stated, we use mini-batches of 64 sentences. We restrict the maximum number of words in a mini-batch to make sure that batches with long sentences

---

backtranslations/en-ro.
[3] http://nlp.stanford.edu/projects/nmt

still fit in GPU memory. If the threshold is exceeded, we simply split the batch until the threshold is met and process the parts separately. Gradients are normalized by the number of non-padding tokens per mini-batch. We also use weight normalization for all layers except for lookup tables (Salimans & Kingma, 2016).

Besides dropout on the embeddings and the decoder output, we also apply dropout to the input of the convolutional blocks (Srivastava et al., 2014). All models are implemented in Torch (Collobert et al., 2011) and trained on a single Nvidia M40 GPU except for WMT'14 English-French for which we use a multi-GPU setup on a single machine. We train on up to eight GPUs synchronously by maintaining copies of the model on each card and split the batch so that each worker computes 1/8-th of the gradients; at the end we sum the gradients via Nvidia NCCL.

### 4.3. Evaluation

We report average results over three runs of each model, where each differs only in the initial random seed. Translations are generated by a beam search and we normalize log-likelihood scores by sentence length. We use a beam of width 5. We divide the log-likelihoods of the final hypothesis in beam search by their length $|\mathbf{y}|$. For WMT'14 English-German we tune a length normalization constant on a separate development set (*newstest2015*) and we normalize log-likelihoods by $|\mathbf{y}|^\alpha$ (Wu et al., 2016). On other datasets we did not find any benefit with length normalization.

For word-based models, we perform unknown word replacement based on attention scores after generation (Jean et al., 2015). Unknown words are replaced by looking up the source word with the maximum attention score in a pre-computed dictionary. If the dictionary contains no translation, then we simply copy the source word. Dictionaries were extracted from the word aligned training data that we obtained with fast_align (Dyer et al., 2013). Each source word is mapped to the target word it is most frequently aligned to. In our multi-step attention (§3.3) we simply average the attention scores over all layers. Finally, we compute case-sensitive tokenized BLEU, except for WMT'16 English-Romanian where we use detokenized BLEU to be comparable with Sennrich et al. (2016b).[4]

---

[4] https://github.com/moses-smt/mosesdecoder/blob/617e8c8/scripts/generic/{multi-bleu.perl,mteval-v13a.pl}

# 5. Results

## 5.1. Recurrent vs. Convolutional Models

We first evaluate our convolutional model on three translation tasks. On WMT'16 English-Romanian translation we compare to Sennrich et al. (2016b) which is the winning entry on this language pair at WMT'16 (Bojar et al., 2016). Their model implements the attention-based sequence to sequence architecture Bahdanau et al. (2014) and uses GRU cells both in the encoder and decoder. We test both word-based and BPE vocabularies (§4).

Table 1 shows that our fully convolutional sequence to sequence model (ConvS2S) outperforms the WMT'16 winning entry for English-Romanian by 1.9 BLEU with a BPE encoding and by 1.3 BLEU with a word factored vocabulary. This instance of our architecture has 20 layers in the encoder and 20 layers in the decoder, both using kernels of width 3 and hidden size 512 throughout. Training took between 6 and 7.5 days on a single GPU.

On WMT'14 English to German translation we compare to the following prior work: Luong et al. (2015) is based on a four layer LSTM attention model, ByteNet (Kalchbrenner et al., 2016) propose a convolutional model based on characters without attention, with 30 layers in the encoder and 30 layers in the decoder, GNMT (Wu et al., 2016) represents the state of the art on this dataset and they use eight encoder LSTMs as well as eight decoder LSTMs, we quote their result for a word-based model, such as ours, as well as a word-piece model (Schuster & Nakajima, 2012).[5]

The results (Table 1) show that our convolutional model outpeforms GNMT by 0.5 BLEU. Our encoder has 15 layers and the decoder has 15 layers, both with 512 hidden units in the first ten layers and 768 units in the subsequent three layers, all using kernel width 3. The final two layers have 2048 units which are just linear mappings with a single input. We trained this model on a single GPU over a period of 18.5 days with a batch size of 48. LSTM sparse mixtures have shown strong accuracy at 26.03 BLEU for a single run (Shazeer et al., 2016) which compares to 25.39 BLEU for our best run. This mixture sums the output of four experts, not unlike an ensemble which sums the output of multiple networks. ConvS2S also benefits from ensembling (§5.2), therefore mixtures are a promising direction.

Finally, we train on the much larger WMT'14 English-French task where we compare to the state of the art result of GNMT (Wu et al., 2016). Our model is trained with a simple token-level likelihood objective and we improve over GNMT in the same setting by 1.6 BLEU on average. We also outperform their reinforcement (RL) models by 0.5

| WMT'16 English-Romanian | BLEU |
| --- | --- |
| Sennrich et al. (2016b) GRU (BPE 90K) | 28.1 |
| ConvS2S (Word 80K) | 29.45 |
| ConvS2S (BPE 40K) | 30.02 |

| WMT'14 English-German | BLEU |
| --- | --- |
| Luong et al. (2015) LSTM (Word 50K) | 20.9 |
| Kalchbrenner et al. (2016) ByteNet (Char) | 23.75 |
| Wu et al. (2016) GNMT (Word 80K) | 23.12 |
| Wu et al. (2016) GNMT (Word pieces) | 24.61 |
| ConvS2S (BPE 40K) | 25.16 |

| WMT'14 English-French | BLEU |
| --- | --- |
| Wu et al. (2016) GNMT (Word 80K) | 37.90 |
| Wu et al. (2016) GNMT (Word pieces) | 38.95 |
| Wu et al. (2016) GNMT (Word pieces) + RL | 39.92 |
| ConvS2S (BPE 40K) | 40.51 |

*Table 1.* Accuracy on WMT tasks comapred to previous work. ConvS2S and GNMT results are averaged over several runs.

BLEU. Reinforcement learning is equally applicable to our architecture and we believe that it would further improve our results.

The ConvS2S model for this experiment uses 15 layers in the encoder and 15 layers in the decoder, both with 512 hidden units in the first five layers, 768 units in the subsequent four layers, 1024 units in the next 3 layers, all using kernel width 3; the final two layers have 2048 units and 4096 units each but the they are linear mappings with kernel width 1. This model has an effective context size of only 25 words, beyond which it cannot access any information on the target size. Our results are based on training with 8 GPUs for about 37 days and batch size 32 on each worker.[6] The same configuration as for WMT'14 English-German achieves 39.41 BLEU in two weeks on this dataset in an eight GPU setup.

Zhou et al. (2016) report a non-averaged result of 39.2 BLEU. More recently, Ha et al. (2016) showed that one can generate weights with one LSTM for another LSTM. This approach achieves 40.03 BLEU but the result is not averaged. Shazeer et al. (2016) compares at 40.56 BLEU to our best single run of 40.70 BLEU.

---

[5]We did not use the exact same vocabulary size because word pieces and BPE estimate the vocabulary differently.

[6]This is half of the GPU time consumed by a basic model of Wu et al. (2016) who use 96 GPUs for 6 days. We expect the time to train our model to decrease substantially in a multi-machine setup.

| **WMT'14 English-German** | **BLEU** |
|---|---|
| Wu et al. (2016) GNMT | 26.20 |
| Wu et al. (2016) GNMT + RL | 26.30 |
| ConvS2S | 26.43 |

| **WMT'14 English-French** | **BLEU** |
|---|---|
| Zhou et al. (2016) | 40.4 |
| Wu et al. (2016) GNMT | 40.35 |
| Wu et al. (2016) GNMT + RL | 41.16 |
| ConvS2S | 41.44 |
| ConvS2S (10 models) | 41.62 |

*Table 2.* Accuracy of ensembles with eight models. We show both likelihood and Reinforce (RL) results for GNMT; Zhou et al. (2016) and ConvS2 use simple likelihood training.

The translations produced by our models often match the length of the references, particularly for the large WMT'14 English-French task, or are very close for small to medium data sets such as WMT'14 English-German or WMT'16 English-Romanian.

### 5.2. Ensemble Results

Next, we ensemble eight likelihood-trained models for both WMT'14 English-German and WMT'14 English-French and compare to previous work which also reported ensemble results. For the former, we also show the result when ensembling 10 models. Table 2 shows that we outperform the best current ensembles on both datasets.

### 5.3. Generation Speed

Next, we evaluate the inference speed of our architecture on the development set of the WMT'14 English-French task which is the concatenation of newstest2012 and newstest2013; it comprises 6003 sentences. We measure generation speed both on GPU and CPU hardware. Specifically, we measure GPU speed on three generations of Nvidia cards: a GTX-1080ti, an M40 as well as an older K40 card. CPU timings are measured on one host with 48 hyperthreaded cores (Intel Xeon E5-2680 @ 2.50GHz) with 40 workers. In all settings, we batch up to 128 sentences, composing batches with sentences of equal length. Note that the majority of batches is smaller because of the small size of the development set. We experiment with beams of size 5 as well as greedy search, i.e beam of size 1. To make generation fast, we do not recompute convolution states that have not been computed compared to the previous time step but rather copy (shift) these activations.

We compare to results reported in Wu et al. (2016) who use Nvidia K80 GPUs which are essentially two K40s. We did not have such a GPU available and therefore run experiments on an older K40 card which is inferior to a K80, in addition to the newer M40 and GTX-1080ti cards. The results (Table 3) show that our model can generate translations on a K40 GPU at 9.3 times the speed and 2.25 higher BLEU; on an M40 the speed-up is up to 13.7 times and on a GTX-1080ti card the speed is 21.3 times faster. A larger beam of size 5 decreases speed but gives better BLEU.

| | **BLEU** | **Time (s)** |
|---|---|---|
| GNMT GPU (K80) | 31.20 | 3,028 |
| GNMT CPU 88 cores | 31.20 | 1,322 |
| GNMT TPU | 31.21 | 384 |
| ConvS2S GPU (K40) $b = 1$ | 33.45 | 327 |
| ConvS2S GPU (M40) $b = 1$ | 33.45 | 221 |
| ConvS2S GPU (GTX-1080ti) $b = 1$ | 33.45 | 142 |
| ConvS2S CPU 48 cores $b = 1$ | 33.45 | 142 |
| ConvS2S GPU (K40) $b = 5$ | 34.10 | 587 |
| ConvS2S CPU 48 cores $b = 5$ | 34.10 | 482 |
| ConvS2S GPU (M40) $b = 5$ | 34.10 | 406 |
| ConvS2S GPU (GTX-1080ti) $b = 5$ | 34.10 | 256 |

*Table 3.* CPU and GPU generation speed in seconds on the development set of WMT'14 English-French. We show results for different beam sizes $b$. GNMT figures are taken from Wu et al. (2016). CPU speeds are not directly comparable because Wu et al. (2016) use a 88 core machine versus our 48 core setup.

On CPU, our model is up to 9.3 times faster, however, the GNMT CPU results were obtained with an 88 core machine whereas our results were obtained with just over half the number of cores. On a per CPU core basis, our model is 17 times faster at a better BLEU. Finally, our CPU speed is 2.7 times higher than GNMT on a custom TPU chip which shows that high speed can be achieved on commodity hardware. We do not report TPU figures as we do not have access to this hardware.

### 5.4. Position Embeddings

In the following sections, we analyze the design choices in our architecture. The remaining results in this paper are based on the WMT'14 English-German task with 13 encoder layers at kernel size 3 and 5 decoder layers at kernel size 5. We use a target vocabulary of 160K words as well as vocabulary selection (Mi et al., 2016; L'Hostis et al., 2016) to decrease the size of the output layer which speeds up training and testing. The average vocabulary size for each training batch is about 20K target words. All figures are averaged over three runs (§4) and BLEU is reported on newstest2014 before unknown word replacement.

We start with an experiment that removes the position em-beddings from the encoder and decoder (§3.1). These embeddings allow our model to identify which portion of the source and target sequence it is dealing with but also impose a restriction on the maximum sentence length. Table 4 shows that position embeddings are helpful but that our model still performs well without them. Removing the source position embeddings results in a larger accuracy decrease than target position embeddings. However, removing both source and target positions decreases accuracy only by 0.5 BLEU. We had assumed that the model would not be able to calibrate the length of the output sequences very well without explicit position information, however, the output lengths of models without position embeddings closely matches models with position information. This indicates that the models can learn relative position information within the contexts visible to the encoder and decoder networks which can observe up to 27 and 25 words respectively.

|  | PPL | BLEU |
|---|---|---|
| ConvS2S | 6.64 | 21.7 |
| -source position | 6.69 | 21.3 |
| -target position | 6.63 | 21.5 |
| -source & target position | 6.68 | 21.2 |

*Table 4.* Effect of removing position embeddings from our model in terms of validation perplexity (valid PPL) and BLEU.

| Attn Layers | PPL | BLEU |
|---|---|---|
| 1,2,3,4,5 | 6.65 | 21.63 |
| 1,2,3,4 | 6.70 | 21.54 |
| 1,2,3 | 6.95 | 21.36 |
| 1,2 | 6.92 | 21.47 |
| 1,3,5 | 6.97 | 21.10 |
| 1 | 7.15 | 21.26 |
| 2 | 7.09 | 21.30 |
| 3 | 7.11 | 21.19 |
| 4 | 7.19 | 21.31 |
| 5 | 7.66 | 20.24 |

*Table 5.* Multi-step attention in all five decoder layers or fewer layers in terms of validation perplexity (PPL) and test BLEU.

Recurrent models typically do not use explicit position embeddings since they can learn where they are in the sequence through the recurrent hidden state computation. In our setting, the use of position embeddings requires only a simple addition to the input word embeddings which is a negligible overhead.

### 5.5. Multi-step Attention

The multiple attention mechanism (§3.3) computes a separate source context vector for each decoder layer. The computation also takes into account contexts computed for preceding decoder layers of the current time step as well as previous time steps that are within the receptive field of the decoder. How does multiple attention compare to attention in fewer layers or even only in a single layer as is usual? Table 5 shows that attention in all decoder layers achieves the best validation perplexity (PPL). Furthermore, removing more and more attention layers decreases accuracy, both in terms of BLEU as well as PPL.

The computational overhead for attention is very small compared to the rest of the network. Training with attention in all five decoder layers processes 3624 target words per second on average on a single GPU, compared to 3772 words per second for attention in a single layer. This is only a 4% slow down when adding 4 attention modules. Most neural machine translation systems only use a single module. This demonstrates that attention is not the bottleneck in neural machine translation, even though it is quadratic in the sequence length (cf. Kalchbrenner et al., 2016). Part of the reason for the low impact on speed is that we batch the computation of an attention module over all target words, similar to Kalchbrenner et al. (2016). However, for RNNs batching of the attention may be less effective because of the dependence on the previous time step.

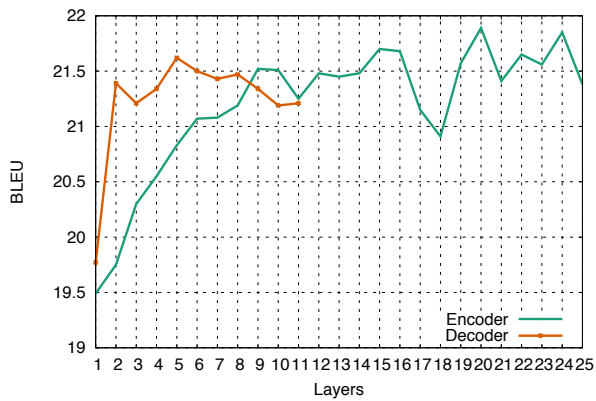

*Figure 2.* Encoder and decoder with different number of layers.

### 5.6. Kernel size and Depth

Figure 2 shows accuracy when we change the number of layers in the encoder or decoder. The kernel width for layers in the encoder is 3 and for the decoder it is 5. Deeper architectures are particularly beneficial for the encoder but less so for the decoder. Decoder setups with two layers already perform well whereas for the encoder accuracy keeps increasing steadily with more layers until up to 9 layers when accuracy starts to plateau.

|  | **DUC-2004** | | | **Gigaword** | | |
|---|---|---|---|---|---|---|
|  | RG-1 (R) | RG-2 (R) | RG-L (R) | RG-1 (F) | RG-2 (F) | RG-L (F) |
| RNN MLE (Shen et al., 2016) | 24.92 | 8.60 | 22.25 | 32.67 | 15.23 | 30.56 |
| RNN MRT (Shen et al., 2016) | 30.41 | 10.87 | 26.79 | 36.54 | 16.59 | 33.44 |
| WFE (Suzuki & Nagata, 2017) | 32.28 | 10.54 | 27.80 | 36.30 | 17.31 | 33.88 |
| ConvS2S | 30.44 | 10.84 | 26.90 | 35.88 | 17.48 | 33.29 |

*Table 6.* Accuracy on two summarization tasks in terms of Rouge-1 (RG-1), Rouge-2 (RG-2), and Rouge-L (RG-L).

| Kernel width | Encoder layers | | |
|---|---|---|---|
|  | 5 | 9 | 13 |
| 3 | 20.61 | 21.17 | 21.63 |
| 5 | 20.80 | 21.02 | 21.42 |
| 7 | 20.81 | 21.30 | 21.09 |

*Table 7.* Encoder with different kernel width in terms of BLEU.

| Kernel width | Decoder layers | | |
|---|---|---|---|
|  | 3 | 5 | 7 |
| 3 | 21.10 | 21.71 | 21.62 |
| 5 | 21.09 | 21.63 | 21.24 |
| 7 | 21.40 | 21.31 | 21.33 |

*Table 8.* Decoder with different kernel width in terms of BLEU.

Aside from increasing the depth of the networks, we can also change the kernel width. Table 7 shows that encoders with narrow kernels and many layers perform better than wider kernels. These networks can also be faster since the amount of work to compute a kernel operating over 3 input elements is less than half compared to kernels over 7 elements. We see a similar picture for decoder networks with large kernel sizes (Table 8). Dauphin et al. (2016) shows that context sizes of 20 words are often sufficient to achieve very good accuracy on language modeling for English.

### 5.7. Summarization

Finally, we evaluate our model on abstractive sentence summarization which takes a long sentence as input and outputs a shortened version. The current best models on this task are recurrent neural networks which either optimize the evaluation metric (Shen et al., 2016) or address specific problems of summarization such as avoiding repeated generations (Suzuki & Nagata, 2017). We use standard likelihood training for our model and a simple model with six layers in the encoder and decoder each, hidden size 256, batch size 128, and we trained on a single GPU in one night. Table 6 shows that our likelihood trained model outperforms the likelihood trained model (RNN MLE) of Shen et al. (2016) and is not far behind the best models on this task which benefit from task-specific optimization and model structure. We expect our model to benefit from these improvements as well.

## 6. Conclusion and Future Work

We introduce the first fully convolutional model for sequence to sequence learning that outperforms strong recurrent models on very large benchmark datasets at an order of magnitude faster speed. Compared to recurrent networks, our convolutional approach allows to discover compositional structure in the sequences more easily since representations are built hierarchically. Our model relies on gating and performs multiple attention steps.

We achieve a new state of the art on several public translation benchmark data sets. On the WMT'16 English-Romanian task we outperform the previous best result by 1.9 BLEU, on WMT'14 English-French translation we improve over the LSTM model of Wu et al. (2016) by 1.6 BLEU in a comparable setting, and on WMT'14 English-German translation we outperform the same model by 0.5 BLEU. In future work, we would like to apply convolutional architectures to other sequence to sequence learning problems which may benefit from learning hierarchical representations as well.

## Acknowledgements

We thank Benjamin Graham for providing a fast 1-D convolution, and Ronan Collobert as well as Yann LeCun for helpful discussions related to this work.

# A. Weight Initialization

We derive a weight initialization scheme tailored to the GLU activation function similar to Glorot & Bengio (2010); He et al. (2015b) by focusing on the variance of activations within the network for both forward and backward passes. We also detail how we modify the weight initialization for dropout.

## A.1. Forward Pass

Assuming that the inputs $\mathbf{x}_l$ of a convolutional layer $l$ and its weights $W_l$ are independent and identically distributed (i.i.d.), the variance of its output, computed as $\mathbf{y}_l = W_l \mathbf{x}_l + \mathbf{b}_l$, is

$$Var[y_l] = n_l Var[w_l x_l] \tag{3}$$

where $n_l$ is the number inputs to the layer. For one-dimensional convolutional layers with kernel width $k$ and input dimension $c$, this is $kc$. We adopt the notation in (He et al., 2015b), i.e. $y_l$, $w_l$ and $x_l$ represent the random variables in $\mathbf{y}_l$, $W_l$ and $\mathbf{x}_l$. With $w_l$ and $x_l$ independent from each other and normally distributed with zero mean, this amounts to

$$Var[y_l] = n_l Var[w_l] Var[x_l]. \tag{4}$$

$\mathbf{x}_l$ is the result of the GLU activation function $\mathbf{y}_{l-1}^a \sigma(\mathbf{y}_{l-1}^b)$ with $\mathbf{y}_{l-1} = (\mathbf{y}_{l-1}^a, \mathbf{y}_{l-1}^b)$, and $\mathbf{y}_{l-1}^a, \mathbf{y}_{l-1}^b$ i.i.d. Next, we formulate upper and lower bounds in order to approximate $Var[x_l]$. If $\mathbf{y}_{l-1}$ follows a symmetric distribution with mean 0, then

$$Var[x_l] = Var[y_{l-1}^a \sigma(\mathbf{y}_{l-1}^b)] \tag{5}$$

$$= E[(y_{l-1}^a \sigma(\mathbf{y}_{l-1}^b))^2] - E^2[y_{l-1}^a \sigma(\mathbf{y}_{l-1}^b)] \tag{6}$$

$$= Var[y_{l-1}^a] E[\sigma(\mathbf{y}_{l-1}^b)^2]. \tag{7}$$

A lower bound is given by $(1/4)Var[y_{l-1}^a]$ when expanding (6) with $E^2[\sigma(y_{l-1}^b)] = 1/4$:

$$Var[x_l] = Var[y_{l-1}^a \sigma(\mathbf{y}_{l-1}^b)] \tag{8}$$

$$= Var[y_{l-1}^a] E^2[\sigma(\mathbf{y}_{l-1}^b)] + \\ Var[y_{l-1}^a] Var[\sigma(\mathbf{y}_{l-1}^b)] \tag{9}$$

$$= \frac{1}{4} Var[y_{l-1}^a] + Var[y_{l-1}^a] Var[\sigma(\mathbf{y}_{l-1}^b)] \tag{10}$$

and $Var[y_{l-1}^a] Var[\sigma(y_{l-1}^b)] > 0$. We utilize the relation $\sigma(x)^2 \le (1/16)x^2 - 1/4 + \sigma(x)$ (Appendix B) to provide an upper bound on $E[\sigma(x)^2]$:

$$E[\sigma(x)^2] \le E\left[\frac{1}{16}x^2 - \frac{1}{4} + \sigma(x)\right] \tag{11}$$

$$= \frac{1}{16}E[x^2] - \frac{1}{4} + E[\sigma(x)] \tag{12}$$

With $x \sim \mathcal{N}(0, std(x))$, this yields

$$E[\sigma(x)^2] \le \frac{1}{16}E[x^2] - \frac{1}{4} + \frac{1}{2} \tag{13}$$

$$= \frac{1}{16}Var[x] + \frac{1}{4}. \tag{14}$$

With (7) and $Var[y_{l-1}^a] = Var[y_{l-1}^b] = Var[y_{l-1}]$, this results in

$$Var[x_l] \le \frac{1}{16}Var[y_{l-1}]^2 + \frac{1}{4}Var[y_{l-1}]. \tag{15}$$

We initialize the embedding matrices in our network with small variances (around 0.01), which allows us to dismiss the quadratic term and approximate the GLU output variance with

$$Var[x_l] \approx \frac{1}{4}Var[y_{l-1}]. \tag{16}$$

If $L$ network layers of equal size and with GLU activations are combined, the variance of the final output $\mathbf{y}_L$ is given by

$$Var[y_L] \approx Var[y_1]\left(\prod_{l=2}^{L}\frac{1}{4}n_l Var[w_l]\right). \tag{17}$$

Following (He et al., 2015b), we aim to satisfy the condition

$$\frac{1}{4}n_l Var[w_l] = 1, \forall l \tag{18}$$

so that the activations in a network are neither exponentially magnified nor reduced. This is achieved by initializing $W_l$ from $\mathcal{N}(0, \sqrt{4/n_l})$.

## A.2. Backward Pass

The gradient of a convolutional layer is computed via back-propagation as $\Delta \mathbf{x}_l = \hat{W}_l \mathbf{y}_l$. Considering separate gradients $\Delta \mathbf{y}_l^a$ and $\Delta \mathbf{y}_l^b$ for GLU, the gradient of $\mathbf{x}$ is given by

$$\Delta \mathbf{x}_l = \hat{W}_l^a \Delta \mathbf{y}_l^a + \hat{W}_l^b \Delta \mathbf{y}_l^b. \tag{19}$$

$\hat{W}$ corresponds to $W$ with re-arranged weights to enable back-propagation. Analogously to the forward pass, $\Delta x_l$, $\hat{w}_l$ and $\Delta y_l$ represent the random variables for the values in $\Delta \mathbf{x}_l$, $\hat{W}_l$ and $\Delta \mathbf{y}_l$, respectively. Note that $W$ and $\hat{W}$ contain the same values, i.e. $\hat{w} = w$. Similar to (3), the variance of $\Delta x_l$ is

$$Var[\Delta x_l] = \hat{n}_l \left( Var[w_l^a] Var[\Delta y_l^a] + Var[w_l^b] Var[\Delta y_l^b] \right). \tag{20}$$

Here, $\hat{n}_l$ is the number of inputs to layer $l+1$. The gradients for the GLU inputs are:

$$\Delta \mathbf{y}_l^a = \Delta \mathbf{x}_{l+1} \sigma(\mathbf{y}_l^b) \quad \text{and} \tag{21}$$

$$\Delta \mathbf{y}_l^b = \Delta \mathbf{x}_{l+1} \mathbf{y}_l^a \sigma'(\mathbf{y}_l^b). \tag{22}$$

The approximation for the forward pass can be used for $Var[\Delta y_l^a]$, and for estimating $Var[\Delta y_l^b]$ we assume an upper bound on $E[\sigma'(y_l^b)^2]$ of $1/16$ since $\sigma'(y_l^b) \in [0, \frac{1}{4}]$. Hence,

$$Var[\Delta y_l^a] - \frac{1}{4} Var[\Delta x_{l+1}] \leq \frac{1}{16} Var[\Delta x_{l+1}] Var[y_l^b]] \tag{23}$$

$$Var[\Delta y_l^b] \leq \frac{1}{16} \Delta Var[\Delta x_{l+1}] Var[y_l^a] \tag{24}$$

We observe relatively small gradients in our network, typically around 0.001 at the start of training. Therefore, we approximate by discarding the quadratic terms above, i.e.

$$Var[\Delta y_l^a] \approx \frac{1}{4} Var[\Delta x_{l+1}] \tag{25}$$

$$Var[\Delta y_l^b] \approx 0 \tag{26}$$

$$Var[\Delta x_l] \approx \frac{1}{4} \hat{n}_l Var[w_l^a] Var[\Delta x_{l+1}] \tag{27}$$

As for the forward pass, the above result can be generalized to backpropagation through many successive layers, resulting in

$$Var[\Delta x_2] \approx Var[\Delta x_{L+1}] \left( \prod_{l=2}^{L} \frac{1}{4} \hat{n}_l Var[w_l^a] \right) \tag{28}$$

and a similar condition, i.e. $(1/4)\hat{n}_l Var[w_l^a] = 1$. In the networks we consider, successions of convolutional layers usually operate on the same number of inputs so that most cases $n_l = \hat{n}_l$. Note that $W_l^b$ is discarded in the approximation; however, for the sake of consistency we use the same initialization for $W_l^a$ and $W_l^b$.

For arbitrarily large variances of network inputs and activations, our approximations are invalid; in that case, the initial values for $W_l^a$ and $W_l^b$ would have to be balanced for the input distribution to be retained. Alternatively, methods that explicitly control the variance in the network, e.g. batch normalization (Ioffe & Szegedy, 2015) or layer normalization (Ba et al., 2016) could be employed.

### A.3. Dropout

Dropout retains activations in a neural network with a probability $p$ and sets them to zero otherwise (Srivastava et al., 2014). It is common practice to scale the retained activations by $1/p$ during training so that the weights of the network do not have to be modified at test time when $p$ is set to 1. In this case, dropout amounts to multiplying activations $\mathbf{x}$ by a Bernoulli random variable $r$ where $\Pr[r = 1/p] = p$ and $\Pr[r = 0] = 1 - p$ (Srivastava et al., 2014). It holds that $E[r] = 1$ and $Var[r] = (1-p)/p$. If $x$ is independent

of $r$ and $E[x] = 0$, the variance after dropout is

$$Var[xr] = E[r]^2 Var[x] + Var[r] Var[x] \tag{29}$$

$$= \left( 1 + \frac{1-p}{p} \right) Var[x] \tag{30}$$

$$= \frac{1}{p} Var[x] \tag{31}$$

Assuming that a the *input* of a convolutional layer has been subject to dropout with a retain probability $p$, the variations of the forward and backward activations from §A.1 and §A.2 can now be approximated with

$$Var[x_{l+1}] \approx \frac{1}{4p} n_l Var[w_l] Var[x_l] \quad \text{and} \tag{32}$$

$$Var[\Delta x_l] \approx \frac{1}{4p} n_l Var[w_l^a] Var[\Delta x_{l+1}]. \tag{33}$$

This amounts to a modified initialization of $W_l$ from a normal distribution with zero mean and a standard deviation of $\sqrt{4p/n}$. For layers without a succeeding GLU activation function, we initialize weights from $\mathcal{N}(0, \sqrt{p/n})$ to calibrate for any immediately preceding dropout application.

## B. Upper Bound on Squared Sigmoid

The sigmoid function $\sigma(x)$ can be expressed as a hyperbolic tangent by using the identity $\tanh(x) = 2\sigma(2x) - 1$. The derivative of $\tanh$ is $\tanh'(x) = 1 - \tanh^2(x)$, and with $\tanh(x) \in [0, 1], x \geq 0$ it holds that

$$\tanh'(x) \leq 1, x \geq 0 \tag{34}$$

$$\int_0^x \tanh'(x)\,dx \leq \int_0^x 1\,dx \tag{35}$$

$$\tanh(x) \leq x, x \geq 0 \tag{36}$$

We can express this relation with $\sigma(x)$ as follows:

$$2\sigma(x) - 1 \leq \frac{1}{2}x, x \geq 0 \tag{37}$$

Both terms of this inequality have rotational symmetry w.r.t 0, and thus

$$\left( 2\sigma(x) - 1 \right)^2 \leq \left( \frac{1}{2}x \right)^2 \quad \forall x \tag{38}$$

$$\Leftrightarrow \sigma(x)^2 \leq \frac{1}{16}x^2 - \frac{1}{4} + \sigma(x). \tag{39}$$

## C. Attention Visualization

Figure 3 shows attention scores for a generated sentence from the WMT'14 English-German task. The model used for this plot has 8 decoder layers and a 80K BPE vocabulary. The attention passes in different decoder layers capture different portions of the source sentence. Layer 1, 3 and 6 exhibit a linear alignment. The first layer shows the clearest alignment, although it is slightly off and frequently attends to the corresponding source word of the previously generated target word. Layer 2 and 8 lack a clear structure and are presumably collecting information about the whole source sentence. The fourth layer shows high alignment scores on nouns such as "festival", "way" and "work" for both the generated target nouns as well as their preceding words. Note that in German, those preceding words depend on gender and object relationship of the respective noun. Finally, the attention scores in layer 5 and 7 focus on "built", which is reordered in the German translation and is moved from the beginning to the very end of the sentence. One interpretation for this is that as generation progresses, the model repeatedly tries to perform the re-ordering. "aufgebaut" can be generated after a noun or pronoun only, which is reflected in the higher scores at positions 2, 5, 8, 11 and 13.

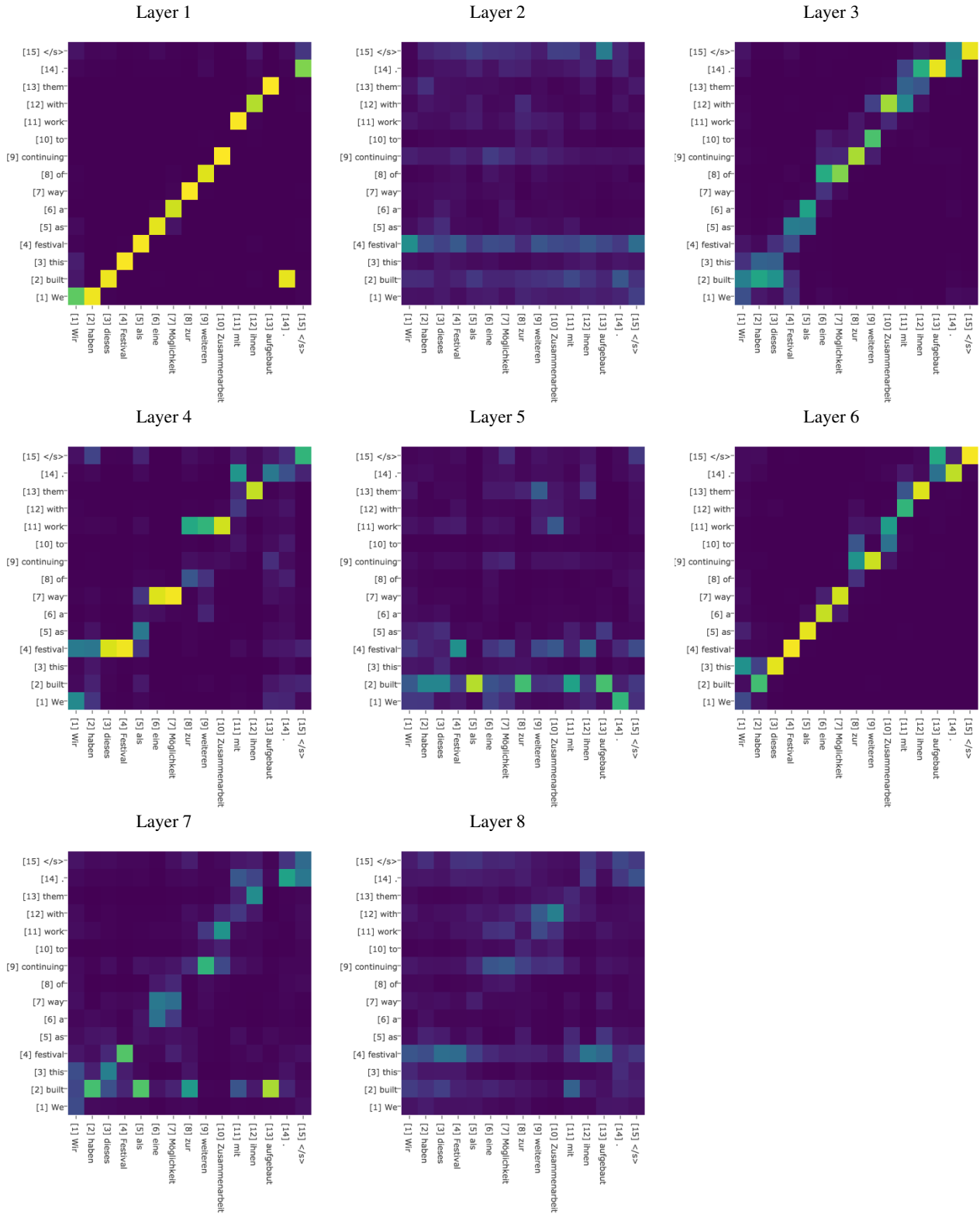

*Figure 3.* Attention scores for different decoder layers for a sentence translated from English (y-axis) to German (x-axis). This model uses 8 decoder layers and a 80k BPE vocabulary.